\documentclass[10pt,twocolumn,letterpaper]{article}

\usepackage{cvpr}              

\usepackage{graphicx}
\usepackage{amsmath}
\usepackage{amssymb}
\usepackage{booktabs}

\usepackage[utf8]{inputenc} 
\usepackage[T1]{fontenc}    
\usepackage{url}            
\usepackage{booktabs}       
\usepackage{amsfonts}       
\usepackage{nicefrac}       
\usepackage{microtype}      
\usepackage{xcolor}         
\usepackage{soul}

\usepackage{times}
\usepackage{epsfig}
\usepackage{graphicx}
\usepackage{amsmath}
\usepackage{amssymb}
\usepackage{multirow}
\usepackage{unicode}

\usepackage{subcaption,booktabs}

\usepackage{adjustbox}

\usepackage{amsfonts}       

\usepackage{times}
\usepackage{epsfig}
\usepackage{graphicx}
\usepackage{amsmath}
\usepackage{amssymb}
\usepackage{multirow}
\usepackage{color}
\usepackage{algorithmic}
\usepackage{algorithm}
\usepackage{booktabs}
\usepackage{url}

\usepackage{colortbl}
\usepackage{bm}
\usepackage{subcaption}
\usepackage{makecell}
\usepackage{pifont}
\usepackage{enumitem}
\usepackage{wrapfig}

\usepackage[title]{appendix}

\usepackage{url}

\newcommand{\cmark}{\ding{51}}%
\usepackage{xcolor}

\usepackage[pagebackref,breaklinks,colorlinks]{hyperref}

\usepackage[capitalize]{cleveref}
\crefname{section}{Sec.}{Secs.}
\Crefname{section}{Section}{Sections}
\Crefname{table}{Table}{Tables}
\crefname{table}{Tab.}{Tabs.}

\def\cvprPaperID{10227} 
\def\confName{CVPR}
\def\confYear{2022}

\begin{document}

\title{SGTR: End-to-end Scene Graph Generation with Transformer}

\author{Rongjie Li\textsuperscript{\rm 1,3,4}
	\quad Songyang Zhang \textsuperscript{\rm 1,3,4}
	\quad Xuming He\textsuperscript{\rm 1,2} \\
	\textsuperscript{\rm 1}School of Information Science and Technology, ShanghaiTech University \quad \\
	\textsuperscript{\rm 2}Shanghai Engineering Research Center of Intelligent Vision and Imaging\\
	\textsuperscript{\rm 3}Shanghai Institute of Microsystem and Information Technology,
	Chinese Academy of Sciences\\
	\textsuperscript{\rm 4}University of Chinese Academy of Sciences\\
	\{lirj2, zhangsy1, hexm\}@shanghaitech.edu.cn
}

\maketitle

\begin{abstract}
  Scene Graph Generation (SGG) remains a challenging visual understanding task due to its compositional property.
  Most previous works adopt a bottom-up two-stage or a point-based one-stage approach, which often suffers from high time complexity or sub-optimal designs.
  In this work, we propose a novel SGG method to address the aforementioned issues, formulating the task as a bipartite graph construction problem. To solve the problem, we develop a transformer-based end-to-end framework that first generates the entity and predicate proposal set, followed by inferring directed edges to form the relation triplets. 
  In particular, we develop a new entity-aware predicate representation based on a structural predicate generator that leverages the compositional property of relationships. Moreover, we design a graph assembling module to infer the connectivity of the bipartite scene graph based on our entity-aware structure, enabling us to generate the scene graph in an end-to-end manner.
  Extensive experimental results show that our design is able to achieve the state-of-the-art or comparable performance on two challenging benchmarks, surpassing most of the existing approaches and enjoying higher efficiency in inference. 
  We hope our model can serve as a strong baseline for the Transformer-based scene graph generation. 
  \footnote{\noindent This work was supported by Shanghai Science and Technology Program 21010502700. 
  Code is available: \url{https://github.com/Scarecrow0/SGTR}
  }

\end{abstract}

\vspace{-0.7cm}
\section{Introduction}\label{sec:intro}
\vspace{-0.2cm}

Inferring structural properties of a scene,
such as the relationship between entities, is a fundamental visual understanding task. The visual relationship between two entities can be typically represented by a triple \textit{ <subject entity, predicate, object entity>}. Based on the visual relationships, a scene can be modeled as a graph structure, with entities as nodes and predicates as edges, referred to as scene graph.
The scene graph provides a compact structural representation for a visual scene, which has potential applications in many vision tasks such as visual question answering~\cite{teney2017graph, shi2019explainable, hildebrandt2020scene}, image captioning~\cite{yang2019auto, yang2021reformer} and image retrieval~\cite{johnson2015image}.


Different from the traditional vision tasks (\textit{e.g.,} object detection) that focus on entity instances, the main challenge of scene graph generation (SGG) lies in building an effective and efficient model for the relations between the entities. 
The compositional property of visual relationships induces high complexity in terms of their constituents, which makes it difficult to learn a compact representation of the relationship concept for localization and/or classification.

\begin{figure}
    \centering
    \includegraphics[width=\linewidth]{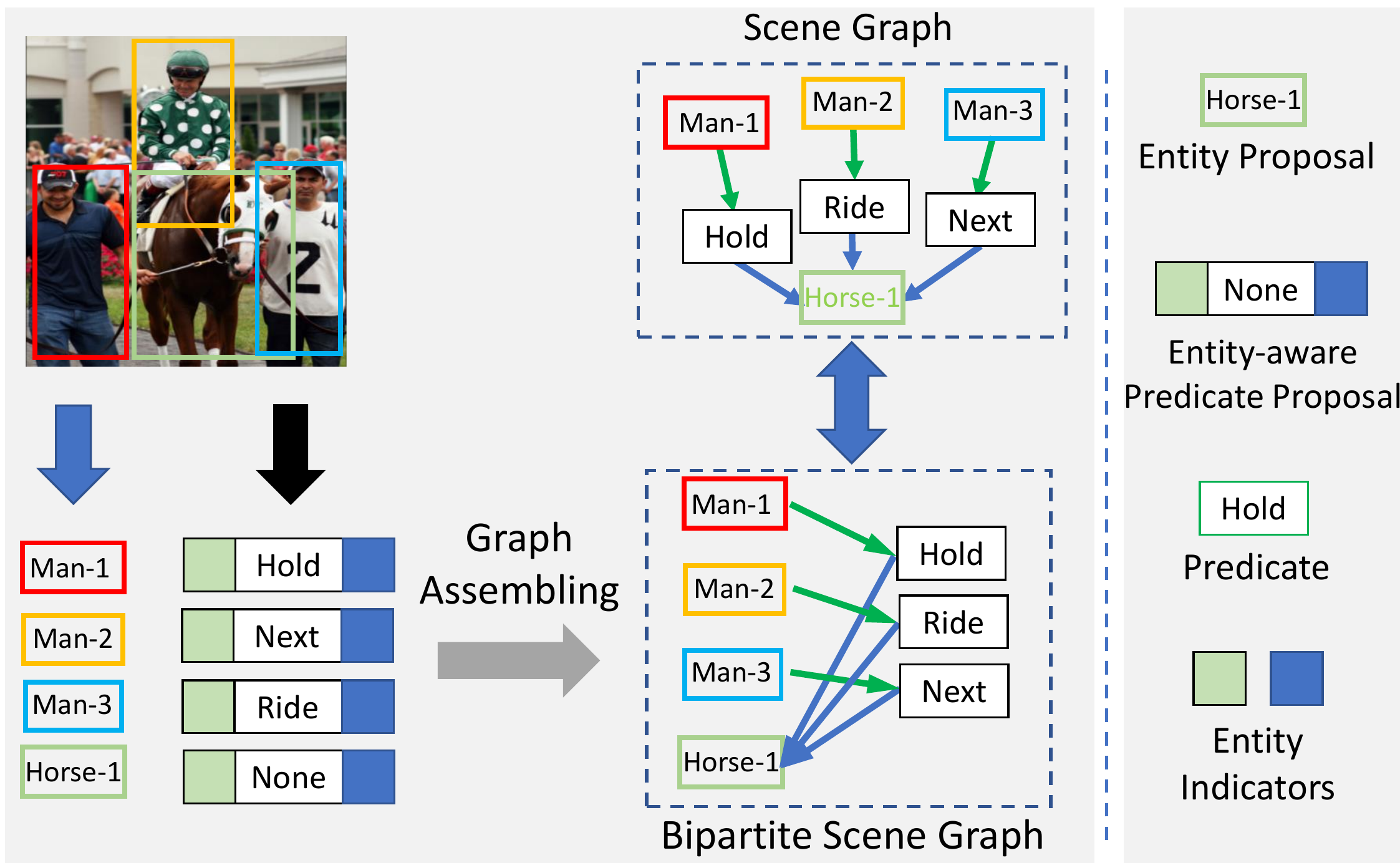}
    \vspace{-0.63cm}
    \caption{\textbf{The illustration of SGTR pipeline paradigm.}
    We formulate SGG as a bipartite graph construction process. First, the entity and predicate nodes are generated, respectively.
    Then we assemble the bipartite scene graph from two types of nodes.
     }
    \label{fig:idea}
    \vspace{-0.65cm}
\end{figure}

Most previous works attempt to tackle this problem using two distinct design patterns: \textit{bottom-up two-stage}~\cite{li2021bipartite, yang2021probabilistic, yao2021visual, desai2021learning, chiou2021recovering, guo2021general, knyazev2021generative, abdelkarim2021exploring} and \textit{point-based one-stage design}~\cite{liu2021fully, dong2021visual}. 
The former typically first detects $N$ entity proposals, followed by predicting the predicate categories of those entity combinations. While this strategy achieves high recalls in discovering relation instances, its $\mathcal{O}(N^2)$ predicate proposals not only incur considerable computation cost but also produce substantial noise in context modeling.
In the one-stage methods, entities and predicates are often extracted separately from the image in order to reduce the size of relation proposal set.  
Nonetheless, they rely on a strong assumption of the non-overlapping property of interaction regions, which severely restricts their application in modeling complex scenes\footnote{
\textit{e.g.,} two different relationships cannot have largely overlapped area -- a phenomenon also discussed in the recent works on (HOI)~\cite{chen2021reformulating, tamura2021qpic}}.

In this work, we aim to tackle the aforementioned limitation by leveraging the compositional property of scene graphs. 
To this end, as illustrated in Fig.~\ref{fig:idea}, we first formulate the SGG task as a bipartite graph construction problem, in which each relationship triplet is represented as two types of nodes (entity and predicate) linked by directed edges. 
Such a bipartite graph allows us to jointly generate entity/predicate proposals and their potential associations, yielding a rich hypothesis space for inferring visual relations. 
More importantly, we propose a novel entity-aware predicate representation that incorporates relevant entity proposal information into each predicate node. 
This enriches the predicate representations and therefore enables us to produce a relatively small number of high-quality predicate proposals. 
Moreover, such a representation encodes potential associations between each predicate and its subject/object entities, which can facilitate predicting the graph edges and lead to efficient generation of the visual relation triplets. 


Specifically, we develop a new transformer-based end-to-end SGG model, dubbed Scene graph Generation TRansformer (SGTR), for constructing the bipartite graph. Our model consists of three main modules, including an \textit{entity node generator}, a \textit{predicate node generator} and a \textit{graph assembling module}.
Given an image, we first introduce two CNN+Transformer sub-networks as the entity and predicate generator to produce a set of entity and predicate nodes, respectively. 
To compute the entity-aware predicate representations, we design a structural predicate generator consisting of three parallel transformer decoders, which fuses the predicate feature with an entity indicator representation.
After generating entity and predicate node representations, we then devise a differentiable \textit{graph assembling} module to infer the directed edges of the bipartite graph, which exploits the entity indicator to predict the best grouping of the entity and predicate nodes.
With end-to-end training, our SGTR learns to infer a sparse set of relation proposals from both input images and entity proposals adaptively, which can mitigate the impact of noisy object detection. 



We validate our method by extensive experiments on two SGG benchmarks: We validate our method by extensive experiments on two SGG benchmarks: Visual Genome and OpenImages-V6 datasets, with comparisons to previous state-of-the-art methods.
The results show that our method outperforms or achieves comparable performance on both benchmarks and with high efficiency during inference.

The main contribution of our work has three-folds: 
\begin{itemize}[itemsep=0mm,topsep=0pt]
    \item We propose a novel transformer-based end-to-end scene graph generation method with a bipartite graph construction process that inherits the advantages of both two-stage and one-stage methods.
    \item We develop an entity-aware structure for exploiting the compositional properties of visual relationships.
    \item Our method achieves the state-of-the-art or comparable performance on all metrics w.r.t the prior SGG methods and with more efficient inference. 
\end{itemize}

\vspace{-0.1cm}
\section{Related Works}
\vspace{-0.1cm}
We categorize the related work of SGG/HOI according to three research directions: \textit{Two-stage Scene Graph Generation}, \textit{One-stage Scene Graph Generation}, and \textit{One-stage Human-Object Interaction}.

\vspace{0.1cm}
\noindent\textbf{Two-stage Scene Graph Generation}~
Two-stage SGG methods predict the relationships between densely connected entity pairs. Based on dense relationship proposals, many previous works focus on modeling contextual structure~\cite{zellers_neural_2017, xu_scene_2017, li_scene_2017, woo_linknet:_2018, tang_learning_2018, li_factorizable_2018, yang_graph_2018, qi_attentive_2018, yin_zoom-net:_2018, wang_exploring_2019, lin_gps-net_2020, zareian_bridging_2020, zareian2020weakly,zareian_learning_2020, cong_nodis_nodate, wang2020tackling, khandelwal2021segmentation, li2021bipartite}. 
Recent studies develop logit adjustment and other training strategies to address the long-tail recognition in the SGG task~\cite{tang_unbiased_2020, knyazev_graph_2020, yan_pcpl_2020, wang2020tackling, suhail2021energy, li2021bipartite, yang2021probabilistic, yao2021visual, desai2021learning, chiou2021recovering, guo2021general, knyazev2021generative, abdelkarim2021exploring}.
The two-stage design is capable of handling complex scenarios encountered in SGG.

However, as discussed in Sec.~\ref{sec:intro}, the dense relation proposal generation often leads to high time complexity and unavoidable noise in context modeling.
Many two-stage works propose heuristic designs to address these issues~(\textit{e.g.}, proposal generation~\cite{yang_graph_2018}, efficient context modeling~\cite{li_factorizable_2018, tang_learning_2018, qi_attentive_2018, yang2019auto, wang_exploring_2019, li2021bipartite}). 
However, these sophisticated designs often rely on the specific properties of the downstream tasks, which limits the flexibility of their representation learning and is difficult to achieve end-to-end optimization.

\vspace{0.1cm}
\noindent\textbf{One-stage Scene Graph Generation}~
Inspired by the fully convolutional one-stage object detection methods~\cite{tian2019fcos,carion2020end,sun2021sparse}, the SGG community starts to explore the one-stage design.
The fully convolutional network~\cite{liu2021fully, teng2021structured} or CNN-Transformer~\cite{dong2021visual} architecture is used in the one-stage methods to detect the relationship from image features directly.
These one-stage frameworks typically can perform efficiently due to their sparse proposal set.
Nonetheless, without explicit entity modeling, those designs may struggle to capture the complex visual relationships associated with real-world scenarios.
Moreover, the majority of one-stage methods ignore entity-relation consistency as they predict each relationship independently rather than a valid graph structure with consistent node-edge constraint.

\vspace{0.1cm}
\noindent\textbf{One-stage Human-Object Interaction}~
Our work is also related to the Human-Object Interaction (HOI) task.
There has been a recent trend toward studying the one-stage framework for Human-Object Interaction~\cite{liao2020ppdm, kim2020uniondet, wang2020learning, zou2021end, chen2021reformulating, tamura2021qpic, kim2021hotr, zhang2021mining}.
In particular, \cite{chen2021reformulating, kim2021hotr} introduce an intriguing framework based on a dual decoder structure that simultaneously extracts the human, object, and interaction and then groups the components into final triplets.
This decoding-grouping approach provides a divide-and-conquer strategy for detecting the human and interacted object.
Inspired by this design, we propose the bipartite graph construction method in our SGTR for the more general SGG task.
To further improve the association modeling between entity and predicate, we propose a predicate node generator with an entity-aware structure and a graph assembling mechanism.
With such a design, the SGTR is able to handle the complex composition of relationships and achieve strong performance on SGG benchmarks.

\vspace{-0.15cm}
\section{Preliminary}
\vspace{-0.1cm}
 
In this section, we first introduce the problem setting of scene graph generation in Sec.~\ref{subsec:setting}, and then present an overview of our approach in Sec.~\ref{subsec:overview}.

\vspace{-0.1cm}
\subsection{Problem Setting}\label{subsec:setting}
\vspace{-0.1cm}

The task of scene graph generation aims to parse an input into a scene graph $\mathcal{G}_{scene}=\{\mathcal{V}_e,\mathcal{E}_r\}$, where $\mathcal{V}_e$ is the node set denoting noun entities and $\mathcal{E}_r$ is the edge set that represents predicates between pairs of subject and object entities.
Specifically, each entity $v_i\in \mathcal{V}_e$ has a category label from a set of entity classes $\mathcal{C}_e$ and a bounding box depicting its location in the image, while each edge $e_{i\to j} \in \mathcal{E}_r $ between a pair of nodes $v_i$ and $v_j$ is associated with a predicate label from a set of predicate classes $\mathcal{C}_p$ in this task.

One possible way to generate the scene graph $\mathcal{G}_{scene}$ is by extracting the relationship triplet set from the given image. 
In this work, we formulate the relationship triplet generation process as a bipartite graph construction task~\cite{li2021bipartite}. 
Specifically, our graph consists of two groups of nodes $\mathcal{V}_e,\mathcal{V}_p$, which correspond to entity representation and predicate representation, respectively. 
These two groups of nodes are connected by two sets of directed edges $\mathcal{E}_{e\rightarrow p}, \mathcal{E}_{p\rightarrow e}$ representing the direction from the entities to predicates and vice versa. Hence the bipartite graph has a form as $\mathcal{G}_b=\{\mathcal{V}_e,\mathcal{V}_p, \mathcal{E}_{e\rightarrow p}, \mathcal{E}_{p\rightarrow e}\}$.


\vspace{-0.1cm}
\subsection{Model Overview}\label{subsec:overview}
\vspace{-0.1cm}

Our model defines a differentiable function $\mathcal{F}_{sgg}$ that takes an image $\mathbf{I}$ as the input and outputs the bipartite graph $\mathcal{G}_{b}$, denoted as $ \mathcal{G}_{b} = \mathcal{F}_{sgg}(\mathbf{I})$, which allows end-to-end training.
We propose to explicitly model the bipartite graph construction process by leveraging the compositional property of relationships. 
The bipartite graph construction consists of two steps: \textit{a) node~(entity and predicate) generation}, and \textit{b) directed edge connection}.

\begin{figure*}[!]
	\vspace{-0.1cm}
	\centering
	\includegraphics[width=0.95\textwidth]{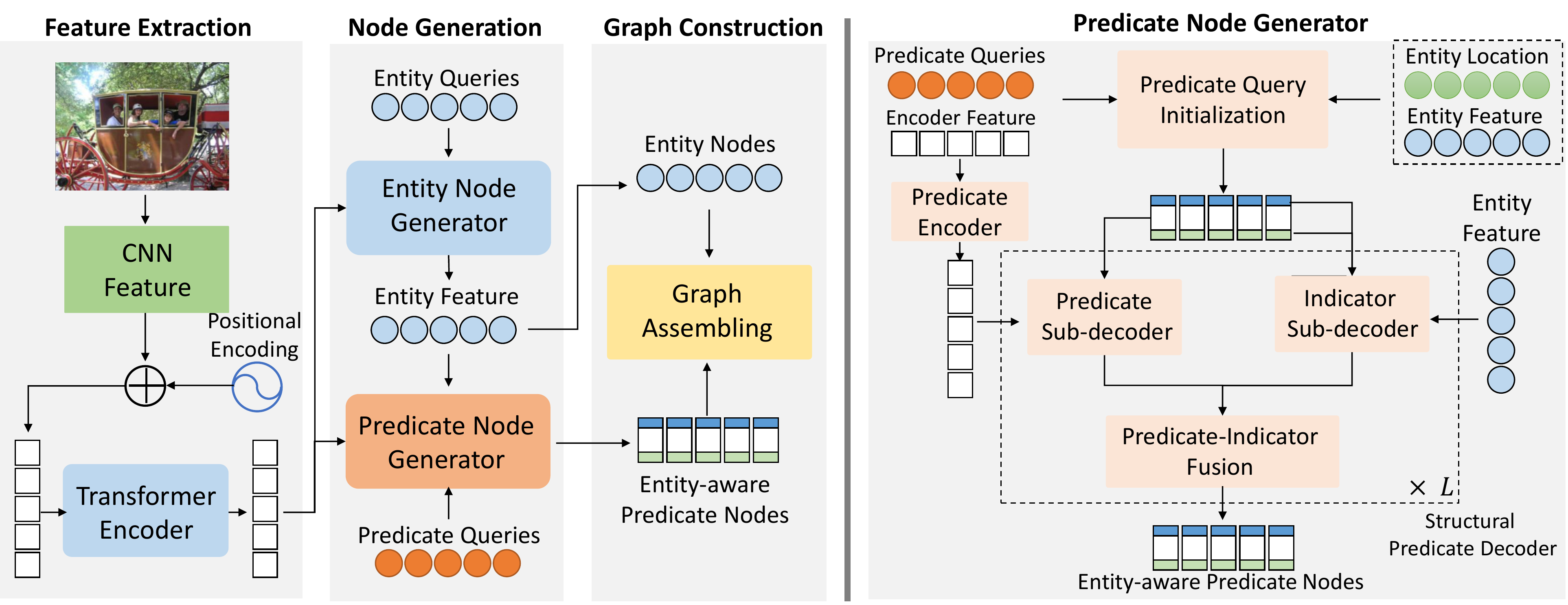}
	\vspace{-0.15cm}
	\caption{
		\textbf{An illustration of overall pipeline of our SGTR model.} \textbf{Left)} We use a CNN backbone together with a transformer encoder for image feature extraction. The entity and predicate node generators are introduced to produce the entity node and entity-aware predicate node. A graph assembling mechanism is developed to construct the final bipartite scene graph. \textbf{Right)} The predicate node generator consists of three parts: \textit{a)} predicate query initialization, \textit{b)} a predicate encoder, and \textit{c)} a structural predicate decoder, which is designed to generate entity-aware predicate nodes. 
	}
	\vspace{-0.6cm}
	\label{fig:main}
\end{figure*}

In the \textit{node generation} step, we extract the entity nodes and predicate nodes from the image with an \textit{entity node generator} and a \textit{predicate node generator}, respectively.
The predicate node generator augments the predicate proposals with entity information based on three parallel sub-decoders.
In the \textit{directed edge connection} step, we design a \textit{graph assembling module} to generate the bipartite scene graph from the entity and predicate proposals.
An overview of our method is illustrated in Fig.~\ref{fig:main} and we will start with a detailed description of our model architecture below.

\vspace{-0.15cm}
\section{Our Approach}\label{subsec:sgtr}
\vspace{-0.15cm}
Our model consists of four main submodules: 
(1) a \textbf{backbone network} for generating feature representation of the scene (Sec.~\ref{subsubsec:backbone}); 
(2) a transformer-based \textbf{entity node generator} for predicting entity proposals (Sec.~\ref{subsubsec:backbone});
(3) a structural \textbf{predicate node generator} for decoding predicate nodes (Sec.~\ref{subsec:triplet});
(4) a \textbf{bipartite graph assembling} module for constructing final bipartite graph via connecting entity nodes and entity-aware predicate nodes (Sec.~\ref{subsec:Assembling}). The model learning and inference are detailed in Sec.~\ref{subsec:learning}.
\vspace{-0.1cm}
\subsection{Backbone and Entity Node Generator}\label{subsubsec:backbone}
\vspace{-0.1cm}

We adopt a ResNet as the backbone network, which first produces a convolutional feature representation for the subsequent modules. 
Motivated by the Transformer-based detector, DETR~\cite{carion2020end}, we then use a multi-layer Transformer encoder to augment the convolutional features. The resulting CNN+transformer feature is denoted as $\mathbf{Z} \in \mathbb{R}^{w\times h \times d}$, where $w,h,d$ are the width, height, and channel of the feature map, respectively.

For the entity node generator, we adopt the decoder of DETR to produce $N_e$ entity nodes from a set of learnable entity queries. Formally, we define the entity decoder as a mapping function $\mathcal{F}_{e}$, which takes initial entity query $\mathbf{Q}_{e}\in \mathbb{R}^{N_e\times d}$ and the feature map $\mathbf{Z}$ as inputs, and outputs the entity locations $\mathbf{B}_e\in{\mathbb{R}^{N_e\times 4}}$ and class scores $\mathbf{P}_e\in{\mathbb{R}^{N_e\times(\mathcal{C}_e+1)}}$, along with their associated feature representations $\mathbf{H}_e\in{\mathbb{R}^{N_e\times d}}$ as follows,
\begin{align}
    \mathbf{B}_e,\mathbf{P}_e,\mathbf{H}_e = \mathcal{F}_{e}(\mathbf{Z}, \mathbf{Q}_{e})
\end{align}
where $\mathbf{B}_e=\{\mathbf{b}_1,\cdots,\mathbf{b}_{N_e}\},\mathbf{b} = (x_c, y_c, w_b, h_b)$, $x_c, y_c$ are the normalized center coordinates of the instance, $w_b,h_b$ are the normalized width and height of each entity box.

\vspace{-0.1cm}
\subsection{Predicate Node Generator}\label{subsec:triplet}
\vspace{-0.1cm}
Our predicate node generator aims to generate an entity-aware predicate representation by incorporating relevant entity proposal information into each predicate node. 
Such a design enables us to encode potential associations between each predicate and its subject/object entities, which can facilitate predicting the graph edges and lead to efficient generation of the visual relation triplets. 

As shown in Fig.~\ref{fig:main}, the predicate node generator is composed of three components: (1) a \textbf{predicate query initialization} module for initializing the entity-aware predicate query (in Sec.~\ref{subsubsec:query_generate}), (2) a \textbf{predicate encoder} for image feature extraction (in Sec.~\ref{subsubsec:triplet_encoder}), and (3) a \textbf{structural predicate decoder} for decoding a set of entity-aware predicate nodes. (in Sec.~\ref{subsubsec:triplet_decoder}).

\vspace{-0.4cm}
\subsubsection{Predicate Encoder}\label{subsubsec:triplet_encoder}
\vspace{-0.2cm}
Based on the CNN+transformer features $\mathbf{Z}$, we introduce a lightweight predicate encoder to extract predicate-specific image features.  
Our predicate encoder, which has a similar structure to the backbone Transformer encoder, employs a form of multi-layer multi-head self-attention via the skip-connected feed-forward network.  
The resulting predicate-specific feature is denoted as $\mathbf{Z}^p\in\mathbb{R}^{w\times h\times d}$.

\vspace{-0.4cm}
\subsubsection{Predicate Query Initialization}\label{subsubsec:query_generate}
\vspace{-0.2cm}
A simple strategy for initializing the predicate queries is to adopt a set of learnable vectors as in the DETR~\cite{carion2020end}. 
However, such a holistic vector-based query design ignores not only the \textit{compositional property} of the visual relationships but also the \textit{entity candidate} information. 
The resulting representations are not expressive enough for capturing the structured and diverse visual relationships.

To cope with this challenge, we introduce a compositional query representation that decouples predicate queries, denoted as $\mathbf{Q}_{p}^e\in \mathbb{R}^{N_r\times 3d}$, into three components $\{ \mathbf{Q}_{is}; \mathbf{Q}_{io}; \mathbf{Q}_{p} \}$, where \textit{subject/object entity indicator}
$\mathbf{Q}_{is},\mathbf{Q}_{io} \in \mathbb{R}^{N_r\times d}$ \footnote{The subscripts 's', 'o' stand for the subject and object entity, respectively. } and \textit{predicate representation} $\mathbf{Q}_{p} \in \mathbb{R}^{N_r\times d}$. 
Concretely, we generate the predicate query $\mathbf{Q}_{p}^e$ in an entity-aware and scene-adaptive manner using a set of initial predicate queries ${\mathbf{Q}}_{init} \in \mathbb{R}^{N_r\times d}$ and entities representation $\mathbf{B}_e, \mathbf{H}_e$. To achieve this, we first build a geometric-aware entity representation as in~\cite{yao2021efficient}, which defines a set of key and value vectors $\in\mathbb{R}^{N_e\times d}$ as follows:
\vspace{-1em}
\begin{align}
\mathbf{K}_{init}=\mathbf{V}_{init}=(\mathbf{H}_e+\mathbf{G}_e), \mathbf{G}_e=\text{ReLU}(\mathbf{B}_e\mathbf{W}_g),
\end{align}
where $\mathbf{G}_e\in\mathbb{R}^{N_e\times d}$ is a learnable geometric embedding of entity proposals, $\mathbf{W}_g\in\mathbb{R}^{4\times d}$ is a transformation from bounding box locations to the embedding space.

Given the augmented entity representations, we then compute the predicate queries $\mathbf{Q}_{p}^e$ using a multi-head cross-attention operation on the initial predicate queries ${\mathbf{Q}}_{init}$ and $\mathbf{K}_{init}$. For clarity, we use $\mathcal{A}(q, k, v)=\text{FFN}(\text{MHA}(q, k, v))$ to denote the multi-head attention operation. 
As such, we have ${\mathbf{Q}}^e_{p}=\mathcal{A}({\mathbf{Q}}_{init},\mathbf{K}_{init},\mathbf{V}_{init}) \mathbf{W}_e$, where $\mathbf{W}_e \in\mathbb{R}^{d\times 3d}=[\mathbf{W}_e^{is},\mathbf{W}_e^{io},\mathbf{W}_e^{p}]$ are the transformation matrices for the three sub-queries~$\mathbf{Q}_{is}, \mathbf{Q}_{io}, \mathbf{Q}_{p}$, respectively. In this way, we obtain a structural query that incorporates the entity information into the predicate query. The sub-queries $\mathbf{Q}_{is},\mathbf{Q}_{io}$ are referred to as entity indicators as they will be used to capture predicate-entity associations below.

\vspace{-0.45cm}
\subsubsection{Structural Predicate Node Decoder}\label{subsubsec:triplet_decoder}
\vspace{-0.15cm}
Given the predicate query $\mathbf{Q}_{q}^{e}$, we now develop a structural predicate node decoder that leverages the compositional property and decodes all the predicate triplets from the entity/predicate feature maps. 

Our structural decoder consists of three modules: a) \textit{predicate sub-decoder}; b) \textit{entity indicator sub-decoders}; c) \textit{predicate indicator fusion}.
The two types of decoders take the encoder feature map $\mathbf{Z}^p$ and entity features $\mathbf{H}_e$, respectively and update the three components of the predicate query independently.
Based on the updated predicate query components, the \textit{predicate-indicator fusion} refines the entire predicate queries, aiming to improve the entity-predicate association within each compositional query. 

Specifically, we adopt the standard transformer decoder structure below. 
For notation clarity, we focus on a single decoder layer and omit layer number $l$ within each sub-decoder, as well as the notation of the self-attention operation. 

\noindent\textbf{Predicate Sub-decoder}.
The predicate sub-decoder is designed to refine the predicate representation from the image feature map $\mathbf{Z}^p$, which utilizes the spatial context in the image for updating predicate representation.
We implement this decoding process using the cross-attention mechanism:~$\widetilde{\mathbf{Q}}_{p}=\mathcal{A}(q=\mathbf{Q}_{p},k=\mathbf{Z}^p,v=\mathbf{Z}^p)$, where
$\widetilde{\mathbf{Q}}_{p}$ is the updated predicate representation.


\noindent\textbf{Entity Indicator Sub-Decoders.}
The entity indicator sub-decoders refine the entity indicators associated with the predicate queries.
Instead of relying on image features, we leverage more accurate entity features in the given scene. Specifically, we perform cross-attention operation between entity indicators
$\mathbf{Q}_{is},\mathbf{Q}_{io}$ and entity proposal features $\mathbf{H}_e$ from the entity node generator,
aiming to enhance the representation of the entity associations.
We denote the updated representation of the entities indicator as $\widetilde{\mathbf{Q}}_{is}, \widetilde{\mathbf{Q}}_{io}$, which are generated with standard cross-attention operation:
\begin{align}
    \widetilde{\mathbf{Q}}_{is}=\mathcal{A}(\mathbf{Q}_{is},\mathbf{H}_e,\mathbf{H}_e),\quad \widetilde{\mathbf{Q}}_{io} =\mathcal{A}(\mathbf{Q}_{io},\mathbf{H}_e,\mathbf{H}_e)
    \vspace{-0.25cm}
\end{align}
\vspace{-1em}

\noindent\textbf{Predicate-Indicator Fusion}\label{para:intr_refine}~
To encode the contextual relation between each predicate query and its entity indicators, we perform a predicate-indicator fusion to calibrate the features of three components in the query.
We explicitly fuse the current $l$-th decoder layer outputs $\widetilde{\mathbf{Q}}_{p}^l, \widetilde{\mathbf{Q}}_{is}^l, \widetilde{\mathbf{Q}}_{io}^l$ to update each component of as the query for next layer~$\mathbf{Q}^{l+1}_{p}, \mathbf{Q}^{l+1}_{is}, \mathbf{Q}^{l+1}_{io}$.
Specifically, we adopt fully connected layers for updating the predicate by fusing entity indicator representations as Eq.~\ref{eq:updatg_pred}:
\vspace{-0.20cm}
\begin{align}
    \mathbf{Q}^{l+1}_{p} = \left(\widetilde{\mathbf{Q}}_{p}^l + \left(\widetilde{\mathbf{Q}}_{is}^l + \widetilde{\mathbf{Q}}_{io}^l\right) \cdot \mathbf{W}_{i}\right) \cdot \mathbf{W}_{p} \label{eq:updatg_pred} 
    \vspace{-0.38cm}
\end{align} 
where $\mathbf{W}_{i}, \mathbf{W}_{p} \in \mathbb{R}^{d \times d}$ are the transformation parameters for update. 
For the entity indicators, we simply adopt the previous layer output as input: $\mathbf{Q}^{l+1}_{is}=\widetilde{\mathbf{Q}}_{is}^l, \mathbf{Q}^{l+1}_{io}=\widetilde{\mathbf{Q}}_{io}^l$.


Based on the refined predicate queries, we are able to generate the geometric and semantic predictions of the predicate node, as well as the location and category of its associated entity indicator as follows,
\begin{align}
    \mathbf{P}_p&=\text{Softmax}(\widetilde{\mathbf{Q}}_{p}\cdot\mathbf{W}_{cls}^{p})\in\mathbb{R}^{N_r\times (\mathcal{C}_p+1)},\\ \mathbf{B}_p&=\sigma(\widetilde{{\mathbf{Q}}}_{p}\cdot\mathbf{W}_{reg}^{p})=\{(x_c^s, y_c^s, x_c^o, y_c^o)\}\in\mathbb{R}^{N_r\times 4}
\end{align}
where $\mathbf{P}_p$ are the class predictions of predicates, and $\mathbf{B}_p=\{(x_c^s, y_c^s, x_c^o, y_c^o)\}$ are the box center coordinates of its subject and object entities. 
The entity indicators are also translated as location prediction of entities $\mathbf{B}_s,\mathbf{B}_o\in\mathbb{R}^{N_r\times 4}$ and their classification predictions $ \mathbf{P}_s,\mathbf{P}_o\in\mathbb{R}^{N_r\times (\mathcal{C}_e+1)}$, which are similar to the entity generator.

Overall, each predicate decoder layer produces the locations and classifications for all the entity-aware predicate queries. Using the multi-layer structure, the predicate decoder is able to gradually improve the quality of predicate and entity association.  

\vspace{-0.1cm}
\subsection{Bipartite Graph Assembling}\label{subsec:Assembling}
\vspace{-0.1cm}
\begin{figure}
    \centering
    \includegraphics[width=0.96\linewidth]{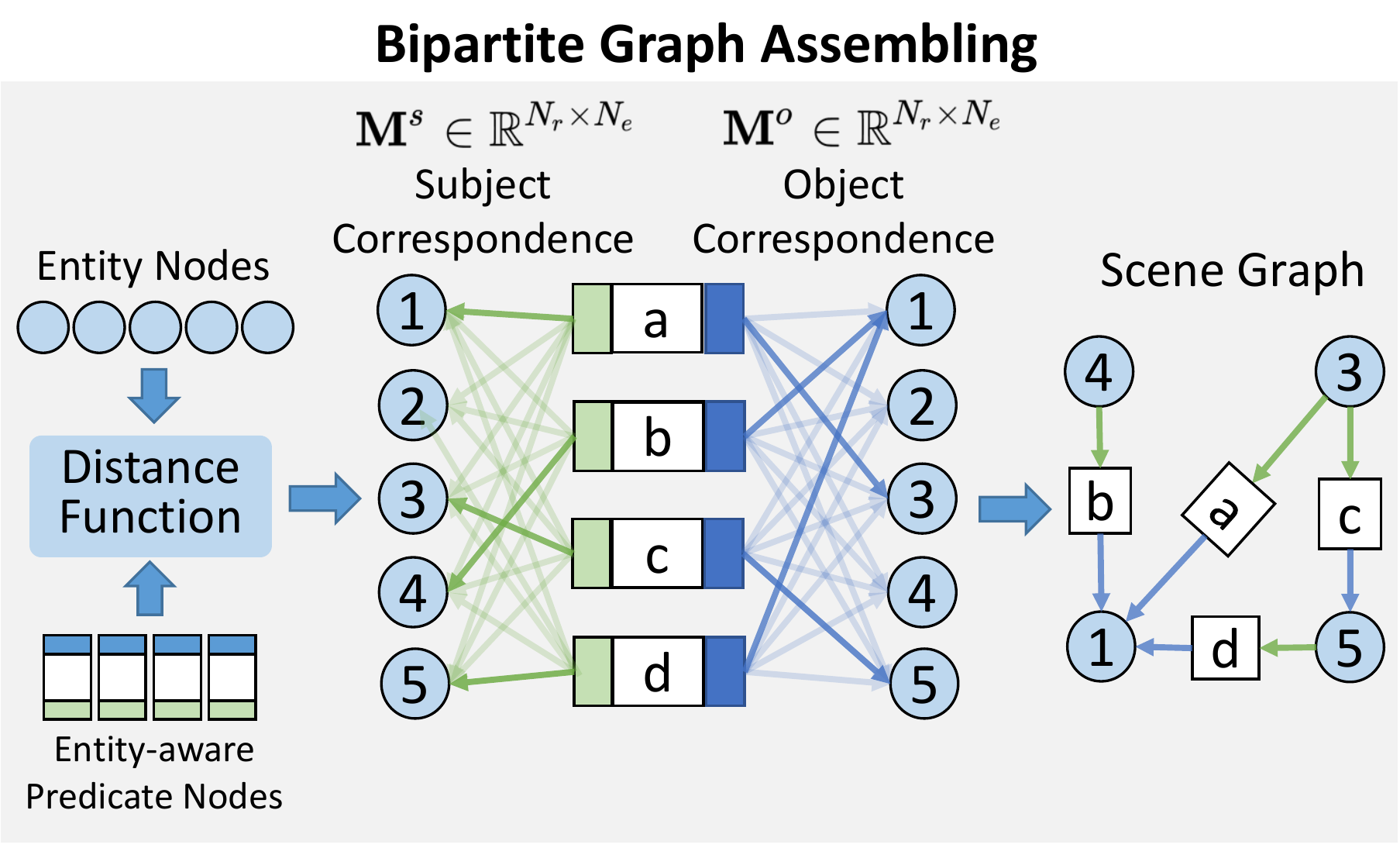}
    \vspace{-0.3cm}
    \caption{\textbf{The illustration of Bipartite Graph Assembling.}}
    \label{fig:modules}
    \vspace{-0.55cm}
\end{figure}

In our formulation, we convert the original scene graph into a bipartite graph structure which consists of $N_e$ entity nodes and $N_r$ predicate nodes, as shown in Fig.~\ref{fig:modules}.
The main goal of the graph assembling is to link the entity-aware predicate nodes to the proper entity node.

To achieve this, we need to obtain the adjacency matrix between the $N_e$ entity nodes and $N_r$ predicate nodes, which can be encoded into a correspondence matrix $\mathbf{M}\in\mathbb{R}^{N_r\times N_e}$. 
Concretely, we define the correspondence matrix by the distance between the entity indicators of predicate nodes and the entity nodes.
Taking the subject entity indicator as example, we have: $\mathbf{M}^s=d_{loc}(\mathbf{B}_s,\mathbf{B}_e) \cdot d_{cls}(\mathbf{P}_s,\mathbf{P}_e)$,
where $d_{loc}(\cdot)$ and $d_{cls}(\cdot)$ are the distance function to measure the matching quality from different dimensions\footnote{\textit{e.g.,} cosine distance between the classification distribution, GIOU and L1 distance between the bounding box predictions, detailed illustration is presented in the supplementary.}.
The correspondence matrix of object entity $\mathbf{M}^o\in\mathbb{R}^{N_r\times N_e}$ is obtained following the same strategy.
The empirical analysis of different distance metrics will be discussed in the experiment section.
Based on the correspondence matrix, we keep the top-$K$ links according to the matching scores as the edge links for each predicate node:
\begin{align}
    \mathbf{R}^s &= \mathcal{F}_{top}(\mathbf{M}^s, K)\in\mathbb{R}^{N_r\times K} \\
    \mathbf{R}^o &= \mathcal{F}_{top}(\mathbf{M}^o, K)\in\mathbb{R}^{N_r\times K}
\end{align}
where $\mathcal{F}_{top}$ is the top-$K$ index selection operation, $\mathbf{R}^s$ and $\mathbf{R}^o$ are the index matrix of entities kept for each triplet from the two relationship roles of subject and object, respectively. 

Using the index matrix $\mathbf{R}^s$ and $\mathbf{R}^o$, we are able to generate the final relationship triplets as $\mathcal{T}=\{(\mathbf{b}_e^s,\mathbf{p}_e^s, \mathbf{b}_{e}^o,\mathbf{p}_e^o, \mathbf{p}_p,\mathbf{b}_p)\}$.
Here $\mathbf{b}_e^s,\mathbf{b}_e^o\in\mathbb{R}^{1\times 4}$ and $\mathbf{p}_e^s,\mathbf{p}_e^o\in\mathbb{R}^{1\times (\mathcal{C}_e+1)}$ are bounding boxes and class predictions of its subject and object entity respectively, 
$\mathbf{p}_p\in\mathbb{R}^{1\times (\mathcal{C}_p+1)}$ is the class prediction of each predicate $\mathbf{P}_p$, and $\mathbf{b}_p\in\mathbf{B}_p$ are the centers of the predicate's entities. 
In the end, the graph assembling module generates the final scene graph as the output of our SGTR model.

\vspace{-0.1cm}
\subsection{Learning and Inference}\label{subsec:learning}
\vspace{-0.1cm}
\noindent\textbf{Learning}~
To train our SGTR model, we design a multi-task loss that consists of two components, including $\mathcal{L}^{enc}$ for the entity generator and $\mathcal{L}^{pre}$ for predicate generator. The overall loss function is formulated as:
\vspace{-0.5em}
\begin{align}
    \mathcal{L} = \mathcal{L}^{enc} + \mathcal{L}^{pre},\quad \mathcal{L}^{pre}=\mathcal{L}^{pre}_{i}+\mathcal{L}^{pre}_{p}
\end{align}
As we adopt a DETR-like detector, the $\mathcal{L}^{enc}$ follows a similar form as \cite{carion2020end}, and the detailed loss equation is reported in the supplementary material. We mainly focus on $\mathcal{L}^{pre}$ in the remaining parts of this section.
To calculate the loss for the predicate node generator, we first obtain the matching matrix between the prediction and the ground truth by adopting the Hungarian matching algorithm~\cite{kuhn1955hungarian}.
We then convert the ground-truth of the visual relationships into a set of triplet representations in as similar form as $\mathcal{T}$, denoted as $\mathcal{T}^{gt}$. The cost of the set matching is defined as:
\vspace{-0.5em}
\begin{align}
    \mathcal{C} = \lambda_{p} \mathcal{C}_p + \lambda_{e} \mathcal{C}_e 
\end{align} 
The two components in the total cost correspond to the costs of predicate and subject/object entity, respectively
\footnote{ 
We utilize the location and classification predictions to calculate cost for each component. Detailed formulations are presented in supplementary.
}.
The matching index $\mathbf{I}^{tri}$ between triplet predictions and ground truths is produced by: $\mathbf{I}^{tri} = \text{argmin}_{\mathcal{T},\mathcal{T}^{gt}} \mathcal{C}$, which is used for following loss calculation of predicate node generator.

The two terms of $\mathcal{L}^{pre}$, that is, $\mathcal{L}^{pre}_{i}, \mathcal{L}^{pre}_{p}$, are used to supervise two types of sub-decoder in predicate node generator. For the entity indicator sub-decoder, we have $\mathcal{L}^{pre}_i=\mathcal{L}^i_{box}+\mathcal{L}^i_{cls}$, where $\mathcal{L}^i_{box}$ and $\mathcal{L}^{i}_{cls}$ are the localization loss (L1 and GIOU loss) and cross-entropy loss for entities indicator $\mathbf{P}_s, \mathbf{B}_s,  \mathbf{P}_o, \mathbf{B}_o$.
Similarly, for the predicate sub-decoder, we have $\mathcal{L}^{pre}_p=\mathcal{L}^p_{ent}+\mathcal{L}^p_{cls}$. 
The $\mathcal{L}^p_{ent}$ is the L1 loss of the location of predicate's associated entities $\mathbf{B}_p$. 
The $\mathcal{L}^{p}_{cls}$ is the cross entropy of predicate category $\mathbf{P}_p$.

\vspace{1mm}
\noindent\textbf{Inference}~
During model inference, we generate $K \cdot N_r$ visual relationship predictions after the assembling stage.
We further remove the invalid self-connection edges during inference.
We adopt a post-processing operation to filter out the self-connected triplets (subject and object entities are identical). 
Then, we rank the remaining predictions by the triplet score $\mathcal{S}_t$ and take the top $N$ relationship triplet as final outputs.
We denote the output as $\mathcal{S}^t = \{(s^t_s \cdot s^t_o \cdot s^t_p)\}$, where $s^t_s, s^t_o$ and $s^t_p$ are the classification probability of subject entity, object entity and predicate, respectively.

\vspace{-0.15cm}
\section{Experiments}
\subsection{Experiments Configuration}
\vspace{-0.15cm}
We evaluate our methods on Openimage V6 datasets~\cite{OpenImages} and Visual Genome~\cite{krishna2017visual}. We mainly adopt the data splits and evaluation metrics from the previous work~\cite{xu_scene_2017,zellers_neural_2017, li2021bipartite}.
For the Openimage benchmark, the weighted evaluation metrics~($\text{wmAP}_{phr}$, $\text{wmAP}_{rel}$, score$_{wtd}$) are used for more class-balanced evaluation.
For the Visual Genome dataset, we adopt the evaluation metric recall@K~(R@K) and mean recall@K~(mR@K) of SGDet, and also report the mR@100 on each long-tail category groups: \textit{head}, \textit{body} and \textit{tail} as same as~\cite{li2021bipartite}.

We use the ResNet-101 and DETR~\cite{carion2020end} as backbone networks and entity detector, respectively.
To speedup training convergence, we first train entity detector on the target dataset, followed by joint training with predicate node generator.
The predicate node generator uses 3 layers of transformer encoder for predicate encoders and 6 layers of transformer decoder for predicate and entity indicator sub-decoders, whose hidden dimensions $d$ is 256. Our predicate decoder uses $N_r$=150 queries. We set  $K$=40 in training and $K$=3 during test for graph assembling module.
For more implementation details please refer to the supplementary.

\vspace{-0.1cm}
\subsection{Ablation Study}
\vspace{-0.1cm}

\begin{table}
    \begin{center}
        \resizebox{\linewidth}{!}{   
            \begin{tabular}{c|ccc|cccc}
                \toprule
               \#  & \textbf{EPN} & \textbf{SPD} & \textbf{GA}  & \textbf{mR@50} & \textbf{mR@100} & \textbf{R@50} & \textbf{R@100}  \\  \midrule
                1 & \cmark & \cmark & \cmark  & \textbf{13.9}  & \textbf{17.3} & \textbf{24.2}   & \textbf{28.2}       \\\midrule
                2 &        & \cmark & \cmark  & 12.0  & 15.9 & 22.9  & 26.3  \\
                3 & \cmark &        & \cmark  & 11.4  & 15.1 & 21.9  & 24.9   \\
                4 &        &        & \cmark  & 11.3  & 14.8 & 21.2  & 24.1   \\
                5 & \cmark & \cmark &         & 4.6   & 7.0  & 10.6   & 13.3     \\ 
                \bottomrule
            \end{tabular}
        }
    \end{center}
    \vspace{-0.5cm}
    \caption{\textbf{Ablation study on model components.} EPN: Entity-aware Predicate Node; SPD: Structural Predicate Decoder, GA: Graph Assembling.} 
    \vspace{-0.5cm}
    \label{comp_abl_table}  
\end{table}

\noindent\textbf{Model Components}~
As shown in Tab.~\ref{comp_abl_table}, we ablate each module to demonstrate the effectiveness of our design on the validation set of Visual Genome.

\noindent$\bullet$ We find that using the holistic query for predicate rather than the proposed structural form decreases the performance by a margin of R@100 and mR@100 at \textbf{1.9} and \textbf{1.4} in line-2.

\noindent$\bullet$ Adopting the shared cross-attention between the image features and predicate/entity indicator instead of the structural predicate decoder leads to the sub-optimal performance as reported in line-3

\noindent$\bullet$ We further remove both entity indicators and directly decode the predicate node from the image feature.
The result is reported in line-4, which decreases the performance by a margin of \textbf{4.2} and \textbf{2.5} on R@100 and mR@100.

\noindent$\bullet$ We also investigate the graph assembling mechanism by directly adopting the prediction of entity indicators as entity nodes for relationship prediction. The poor results shown in line-5 demonstrate that the model struggles to tackle such complex multi-tasks within a single structure, while proposed entity-prediction association modeling and graph assembling reduce the difficulty of optimization.

\begin{table}
    \begin{center}
        \resizebox{0.42\textwidth}{!}{      
            \begin{tabular}{ll|cccc}
                \toprule
                \textbf{NPD} & \textbf{NED} & \textbf{mR@50}& \textbf{mR@100} & \textbf{R@50} & \textbf{R@100}  \\ \midrule
                3    & 3    & 10.6  & 13.3 & 23.4   & 27.4   \\
                6    & 6   & \textbf{13.9}  & \textbf{17.3} & \textbf{24.2}   & 28.2     \\
                12   & 12   & 13.7  & 17.0  & 24.0   & \textbf{28.4}   \\ \bottomrule
                \end{tabular}
            }
    \end{center}
    \vspace{-0.48cm}
    \caption{\textbf{Ablation study on number of predicate decoder layers.} NPD: number of predicate sub-decoder layers; NED: number of entity indicator sub-decoder layers;} 
    \label{model_scale_table} 
    \vspace{-0.3cm}
\end{table}

\begin{table}
    \centering
        \resizebox{0.40\textwidth}{!}{        
            \begin{tabular}{l|cccc}
                \toprule
                \textbf{GA} & \textbf{mR@50} & \textbf{mR@100} & \textbf{R@50} & \textbf{R@100}  \\ \midrule
                S & 10.6    & 11.8   & \textbf{24.4}   & 27.7            \\
                F  &  13.3   &  16.1 &  23.7   &   27.5  \\
                \textbf{Ours} & \textbf{13.9}  & \textbf{17.3}    & 24.2  & \textbf{28.2}   \\ \bottomrule
                \end{tabular}
        }
        \vspace{-0.20cm}
        \caption{\textbf{Ablation study on graph assembling}, S: spatial distance between the predicate and entity-based matching function proposed by AS-Net\cite{chen2021reformulating}; F: feature similarity-based matching function proposed by HOTR~\cite{kim2021hotr}.} 
        \label{ga_abl_table} 
        \vspace{-0.7cm}
\end{table}

\noindent\textbf{Graph Assembling Design}~
We further investigate the effectiveness of our graph assembling design. Specifically, we adopt the differentiable entity-predicate pair matching function proposed by recent HOI methods~\cite{chen2021reformulating, kim2021hotr}, as shown in Tab.~ \ref{ga_abl_table}. 
Comparison experiments are conducted on the validation set of Visual Genome by using different distance functions for the assembling module.
In AS-Net\cite{chen2021reformulating}, the grouping is conducted based on the distance between entity bounding box and entity center predicted by interaction branch, which lacks the entity semantic information. 
The HOTR~\cite{kim2021hotr} introduces a cosine similarity measurement between the predicate and entity in feature space. We implement this form for calculation the distance between the entity indicator $\widetilde{\mathbf{Q}}_{is}$, $\widetilde{\mathbf{Q}}_{io}$ and entity nodes $\mathbf{H}_e$.
Compared with location-only~\cite{chen2021reformulating} similarity and feature-based~\cite{kim2021hotr} similarity, our proposed assembling mechanism, taking both semantic and spatial information into the similarity measurement, is preferable. We also empirically observe that the feature-based~\cite{kim2021hotr} similarity design has a slower and more unstable convergence process.

\noindent\textbf{Model Size}
To investigate the model complexity of the structural predicate node decoder, we incrementally vary the number of layers $L$ in the predicate and entity indicator decoder.
The quantitative results are shown in Tab.~\ref{model_scale_table}.
The results indicate that our model achieves the best performance while $L=6$. 
We observe that the performance improvement is considerable when increasing the number of decoder layers from 3 to 6, and performance will be saturated when $L = 12$.

\noindent\textbf{Entity Detector}
As we adopt different entity detectors compared to previous two-stage designs, we conduct experiments to analyze the influence of detectors on the SGTR. The detailed results are presented in the supplementary.

\vspace{-0.1cm}
\subsection{Comparisons with State-of-the-Art Methods} \label{subsec:sota_comp}
\vspace{-0.1cm}

We conduct experiments on Openimage-V6 benchmark and VG dataset to demonstrate the effectiveness of our design. 
We compare our method with several state-of-the-art two-stage(\textit{e.g.}, VCTree-PCPL, VCTree-DLFE, BGNN~\cite{li2021bipartite}, VCTree-TDE, DT2-ACBS~\cite{desai2021learning}) and one-stage methods(\textit{e.g.} AS-Net, HOTR, FCSGG) on Visual Genome dataset.
Since our backbone is different from what they reported, we reproduced the SOTA methods BGNN and its baseline RelDN with the same ResNet-101 backbone for more fair comparisons.
Furthermore, since FCSGG\cite{liu2021fully} is the only published one-stage method for SGG, we reproduce the result of several strong one-stage HOI methods with similar entity-predicate pairing mechanisms (AS-Net~\cite{chen2021reformulating}, HOTR~\cite{kim2021hotr}) using their released code for a more comprehensive comparison.

\begin{table}
    \vspace{-0.1cm}
    \begin{center}
        \resizebox{0.46\textwidth}{!}{        
            \begin{tabular}{l|l|cc|cc|c}
                \toprule 
            \multirow{2}{*}{\textbf{B}} & \multirow{2}{*}{\textbf{Models}} &\multirow{2}{*}{\textbf{mR@50}} &  \multirow{2}{*}{\textbf{R@50}} &\multicolumn{2}{c|}{\textbf{wmAP}} 
             & \multirow{2}{*}{\textbf{score}\scriptsize{wtd}}    \\ 
             \cmidrule{5-6}
             &  &  &   & \textbf{rel}& \textbf{phr} &    \\ 
                \midrule
                \multirow{3}{*}{\rotatebox{90}{X101-F}} 
                     & RelDN   & 37.20  & \textbf{75.40} & 33.21 & 31.31   & 41.97   \\
                     & GPS-Net & 38.93  & 74.74 & 32.77 & 33.87  & 41.60   \\
                     & BGNN    & 40.45  & 74.98 & 33.51  & 34.15   & 42.06  \\ \midrule
                \multirow{5}{*}{\rotatebox{90}{R101}}     
                & BGNN$^{*\dagger}$  & 39.41 & 74.93  & 31.15 & 31.37 &  40.00  \\
                & RelDN$^{\dagger}$ & 36.80 & 72.75  & 29.87 & 30.42 &  38.67  \\ \cmidrule{2-7}
                & HOTR$^{\dagger}$   & 40.09 & 52.66 & 19.38 & 21.51 &  26.88   \\
                & AS-Net$^{\dagger}$ & 35.16 & 55.28 & 25.93 & 27.49 &  32.42 \\
                & \textbf{Ours}  & \textbf{42.61} & 59.91 & \textbf{36.98} & \textbf{38.73} &  \textbf{42.28}  \\ \bottomrule
                \end{tabular}

        }
    \end{center}
\vspace{-0.5cm}
\caption{\textbf{The Performance on Openimage V6.} 
        $\dagger$ denotes results reproduced with the authors' code. The performance of ResNeXt-101 FPN is borrow from \cite{li2021bipartite}. * means using resampling strategy.} 
        \vspace{-0.7cm}
        \label{oiv6_overall_table} 
\end{table}

\noindent\textbf{OpenImage V6}~
The performance on the OpenImage V6 dataset is reported in Tab.~\ref{oiv6_overall_table}.
We re-implement the SOTA one-stage and two-stage methods with the same ResNet-101 backbone.
Our method outperforms the two-stage SOTA method BGNN with an improvement of \textbf{2.28}.
Specifically, our design has a significant improvement on weighted mAP metrics of relationship detection (${\mathbf{\text{wmAP}}_{rel}}$) and phrase detection (${\mathbf{\text{wmAP}}_{phr}}$) sub-tasks of \textbf{5.83} and \textbf{7.36} respectively, which indicates that leveraging the compositional property of the visual relationship is beneficial for the SGG task.

\begin{table*}[!ht]
    \vspace{-0.1cm}
    \begin{center}
        \resizebox{0.71\textwidth}{!}{        
                \begin{tabular}{l|l|l|cc|ccc|c}
                \toprule
                \textbf{B}   & \textbf{D}    & \textbf{Method}       & \textbf{mR@50/100} & \textbf{R@50/100} & \textbf{Head}  & \textbf{Body}  & \textbf{Tail}  & \textbf{Time/Sec} \\ \midrule
                $\star$ &  $\star$ & FCSGG  \cite{liu2021fully}  & 3.6~/~4.2 & 21.3~/~25.1 &-&-&- & 0.12 \\\midrule
                \multirow{9}{*}{\rotatebox{90}{X101-FPN}} & \multirow{12}{*}{\rotatebox{90}{Faster-RCNN}} 
                                            & RelDN \cite{li2021bipartite}     & 6.0~/~7.3   & 31.4~/~35.9 & - & - & - & 0.65   \\
                                          & & Motifs\cite{tang_unbiased_2020}     & 5.5~/~6.8 &  \textbf{32.1}~/~\textbf{36.9}  & -  & - & - & 1.00   \\
                                          & & VCTree\cite{tang_unbiased_2020}     & 6.6~/~7.7 &  31.8~/~36.1  & -  & - & - & 1.69   \\
                                          & & BGNN$^{*\dagger}$ \cite{li2021bipartite}   & 10.7~/~12.6 & 31.0~/~35.8 & 34.0 & 12.9 &  6.0 & 1.32\\ \cmidrule{3-9} 
                                          & & VCTree-TDE\cite{tang_unbiased_2020}        &  9.3~/~11.1 &  19.4~/~23.2 & - & - & - & $\ge$1.69   \\
                                          & & VCTree-DLFE \cite{chiou2021recovering}         & 11.8~/~13.8 & 22.7~/~26.3 & -& - &- &$\ge$1.69 \\
                                          & & VCTree-EBM \cite{suhail2021energy}          & 9.7~/~11.6 & 20.5~/~24.7 & - & - & - &$\ge$1.69 \\
                                          & & VCTree-BPLSA \cite{guo2021general}          & 13.5~/~15.7 & 21.7~/~25.5 & - & - & - &$\ge$1.69 \\
                                          & & DT2-ACBS \cite{desai2021learning}          & \textbf{22.0}~/~\textbf{24.4} & 15.0~/~16.3 & - & - & - & $\sim$0.63 \\ \cmidrule{1-1} \cmidrule{3-9} 
                \multirow{6}{*}{\rotatebox{90}{R101}}     
                                          && BGNN$^{*\dagger}$ & 8.6~/~10.3 &  28.2~/~33.8 & 29.1 & 12.6 & 2.2 &  1.32 \\
                                          && RelDN$^{\dagger}$ & 4.4~/~5.4  & \textbf{30.3}~/~\textbf{34.8} & \textbf{31.3} & 2.3  & 0.0 & 0.65\\ \cmidrule{2-9} 
                                          & \multirow{4}{*}{\rotatebox{90}{DETR}}        
                                          & AS-Net$^{\dagger}$  \cite{chen2021reformulating} & 6.12~/~7.2   & 18.7~/~21.1 & 19.6 & 7.7 & 2.7 & 0.33   \\
                                          && HOTR$^{\dagger}$ \cite{kim2021hotr}     & 9.4~/~12.0  &  23.5~/~27.7 & 26.1 & 16.2 & 3.4 & 0.25   \\
                                          \cmidrule{3-9}
                                     && \textbf{Ours$^{\diamond}$}    & 12.0~/~14.6  & 25.1~/~26.6  & 27.1 & 17.2 & 6.9 & 0.35      \\
                                          && \textbf{Ours}        & 12.0~/~15.2    & 24.6~/~28.4& 28.2 & 18.6 & 7.1 & 0.35      \\

                                          && \textbf{Ours}$^{*}$  &  \textbf{15.8}~/~\textbf{20.1} & 20.6~/~25.0 & 21.7 & \textbf{21.6} & \textbf{17.1}& 0.35   \\ \bottomrule
                \end{tabular}
        }
    \end{center}
    \vspace{-0.48cm}
    \caption{\textbf{The SGDet performance on test set of Visual Genome dataset.} $\dagger$ denotes results reproduced with the authors' code. $*$ denotes the bi-level resampling \cite{li2021bipartite} is applied for this model. {$\diamond$ denotes that our model uses $K=1$ for top-$K$ matching in graph assembling (more ablative experiments for $K$ are presented in the supplementary)}.
    $\star$ denotes the special backbone  HRNetW48-5S-FPN$_{\times \text{2-f}}$ 
    and entities detector, CenterNet\cite{zhou2019objects}.} 
    \vspace{-0.6cm}
\label{vg_overall_table}
\end{table*}

\noindent\textbf{Visual Genome}~
As shown in Tab. \ref{vg_overall_table}, with the same ResNet-101 backbone,  we compare our method with the two-stage method BGNN~\cite{li2021bipartite}, and the one-stage methods HOTR~\cite{kim2021hotr}, AS-Net~\cite{chen2021reformulating}.
It shows that our method outperforms HOTR with a significant margin of \textbf{4.9} and \textbf{3.2} on mRecall@100.
Furthermore, our method achieves considerable improvement when compared with the two-stage methods, and the detailed  performance is presented in the supplementary.

\noindent$\bullet$ Benefitting from the sparse proposal set, SGTR has a more balanced foreground/background proposal distribution than the traditional two-stage design, where there exists a large number of negative samples due to exhausted entity pairing. 
Thus, when equipped with the same backbone and learning strategy as before, our method achieves competitive performance in mean recall. 
We also list several newly proposed works, which develops various training strategies for long-tailed recognition.
Our method achieves higher mR@100 performance with less overall performance drop when using the resampling strategy proposed in \cite{li2021bipartite}. 
We refer the reader to the supplementary for more experiments on our model using advanced long-tail training strategies.

\noindent$\bullet$ 
We find that the performance of our model in the head category is lower than the two-stage methods with the same backbone. 
The main reason is that the DETR detector performs weaker on small entities than the traditional Faster-RCNN. 
Since the visual genome has a large proportion of relationships involving small objects, our method performs sub-optimal in recognizing those relationships. {The detailed limitation analysis is presented in the supplementary.}

\noindent$\bullet$
We compare the efficiency of SGTR with previous methods according to the inference time (seconds/image) on the NVIDIA GeForce Titan XP GPU with a batch size of 1 and an input size of 600 x 1000.
Our design obtains comparable inference time as the one-stage methods using the same backbone, which demonstrates the efficiency of our method.

\vspace{-0.15cm}
\subsection{Qualitative Results}
\begin{figure}
    \centering
    \includegraphics[width=0.47\textwidth]{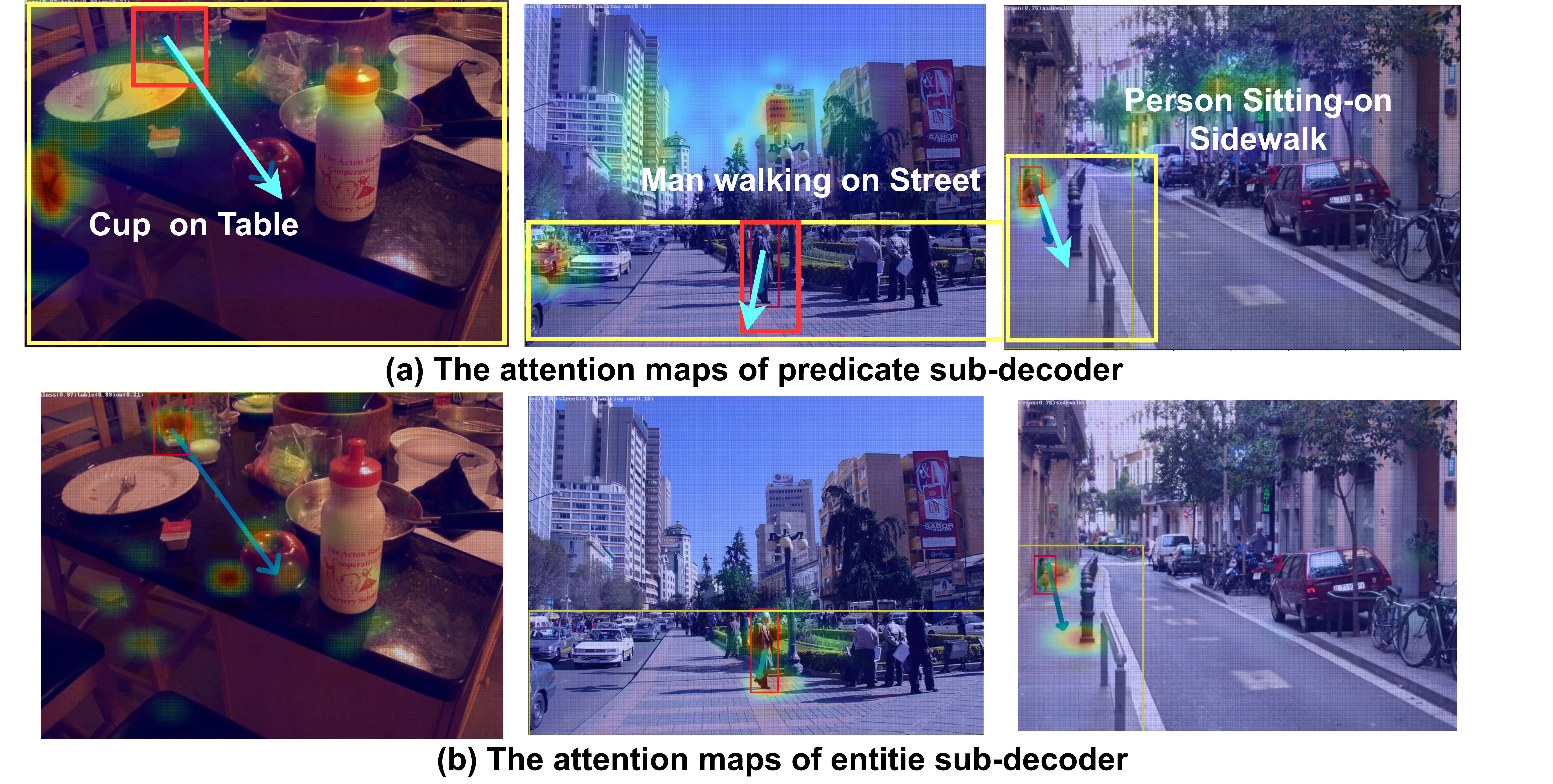}
    \vspace{-0.1cm}
    \caption{\textbf{The visualization on attention heatmap of structural predicate decoder.} The predicate sub-decoder focus on contextual representation around the entities of triplets.
Entity indicator sub-decoders focus on relationship-based entity regions.}
    \vspace{-0.75cm}
    \label{fig:vis}
\end{figure}

\vspace{-0.15cm}
As shown in Fig. \ref{fig:vis}, we visualize the attention weight of the predicates sub-decoder and entity sub-decoder on images from the validation set of the Visual Genome dataset.
By comparing the heatmaps in Fig. \ref{fig:vis} (a) and Fig. \ref{fig:vis} (b),
we note that for the same triplet prediction, the predicate sub-decoder focuses more on the contextual regions around the entities of triplets while the entity sub-decoders put more attention on the entity regions.
Therefore, our design allows the model to learn the compositional property of visual relationships more effectively, which improves prediction accuracy.
More visualization results are reported in the supplementary (including analysis of graph assembling, comparison between two-stage methods, etc.).

\vspace{-1em}
\section{Conclusions}

In this work, we propose a novel end-to-end CNN-Transformer-based scene graph generating approach (SGTR).
In comparison to the prior approaches, our major contribution consists of two components:
We formulate the SGG as a bipartite graph construction with three steps: entity and predicate nodes generation and directed edge connection. 
We develop the entity-aware representation for modeling the predicate nodes, which is integrated with the entity indicators by the structural predicate node decoder. 
Finally, the scene graph is constructed by the graph assembling module in an end-to-end manner.
Extensive experimental results show that our SGTR outperforms or is competitive with previous state-of-the-art methods on the Visual Genome and Openimage V6 datasets.

\noindent{\textbf{Potential Negative Societal Impact} One possible negative impact is that SGG may serve as a base module for surveillance abuse and collecting private information.}



{\small
\bibliographystyle{ieee_fullname}
\bibliography{main}
}



\end{document}